\documentclass[11pt,a4paper,hyphens,]{article}

\author{\hspace{3.2em}{\centering Valerio Perrone,$^{\clubsuit}$\thanks{ ~ Work done prior to joining Amazon.} ~ Simon Hengchen,$^{\spadesuit}$ Marco Palma,$^{\heartsuit}$ Alessandro Vatri,$^{\dagger}$}\\ 
Jim Q. Smith,$^{\heartsuit\diamondsuit}$ Barbara McGillivray$^{\triangle\diamondsuit}$ \\
{\footnotesize $^{\clubsuit}$Amazon ~ $^{\spadesuit}$University of Gothenburg ~ $^{\dagger}$University of Oxford}\\
{\footnotesize $^{\heartsuit}$University of Warwick ~ $^{\diamondsuit}$The Alan Turing Institute ~ $^{\triangle}$University of Cambridge}\\
{{\tt \footnotesize vperrone@amazon.com}}
}

\title{Lexical semantic change for Ancient Greek and Latin}

\usepackage{tabularx}

\usepackage{tgpagella}
\usepackage{fancyhdr}
\usepackage{lastpage}
\usepackage{latexsym}
\usepackage{url}
\usepackage{wrapfig}
\usepackage{multirow}
\usepackage{microtype}
\usepackage{graphicx}
\usepackage{subfigure}
\usepackage{booktabs} 
\usepackage{enumitem,xcolor}
\usepackage[normalem]{ulem}
\usepackage[ruled,linesnumbered]{algorithm2e}
\usepackage{tikz}
\usepackage{natbib}
\usetikzlibrary{fit,positioning, calc}

\tikzset{
    between/.style args={#1 and #2}{
         at = ($(#1)!0.5!(#2)$)
    }
}

\usepackage[hidelinks]{hyperref}       
\hypersetup{
    colorlinks,
    linkcolor={red!50!black},
    citecolor={blue!50!black},
    urlcolor={blue!80!black}
}

\usepackage{url}            
\usepackage{booktabs}       
\usepackage{amsmath}
\usepackage{amsthm}
\usepackage{nicefrac}       
\usepackage{microtype}      
\usepackage{color}
\usepackage{graphicx}
\usepackage{tikz}
\usepackage{mathtools}
\usepackage{comment}
\usetikzlibrary{arrows.meta,calc,decorations.markings,math}
\usetikzlibrary{bayesnet}
\theoremstyle{plain}

\theoremstyle{definition}

\usepackage{listingsutf8}
\usepackage{listings} 
\usepackage{tgpagella}

\begin{document}
\maketitle
\abstract{Change and its precondition, variation, are inherent in languages. Over time, new words enter the lexicon, others become obsolete, and existing words acquire new senses. Associating a word's correct meaning in its historical context is a central challenge in diachronic research. Historical corpora of classical languages, such as Ancient Greek and Latin, typically come with rich metadata, and existing models are limited by their inability to exploit contextual information beyond the document timestamp. While embedding-based methods feature among the current state of the art systems, they are lacking in the interpretative power. In contrast, Bayesian models provide explicit and interpretable representations of semantic change phenomena. In this chapter we build on GASC, a recent computational approach to semantic change based on a dynamic Bayesian mixture model. In this model, the evolution of word senses over time is based not only on distributional information of lexical nature, but also on text genres. We provide a systematic comparison of dynamic Bayesian mixture models for semantic change with state-of-the-art embedding-based models. On top of providing a full description of meaning change over time, we show that Bayesian mixture models are highly competitive approaches to detect binary semantic change in both Ancient Greek and Latin.}

\section{Introduction}

The study of lexical semantics in a diachronic perspective is of primary importance in lexicography, historical linguistics and other humanistic fields. 
Capturing the semantic spectrum and historical change of individual words as well as performing large-scale diachronic analyses of the lexicon  can help us answer important questions about the development of our culture and heritage. 
Recent research in Natural Language Processing (NLP) has led to the development of computational models of lexical semantic change (LSC) which have the potential to add new insights to diachronic semantics. Most computational research in this area, however, has focussed on extant languages, and only a few attempts have been made to tackle this topic for ancient languages. 

To address this, \cite{perrone-etal-2019-gasc} introduced GASC (\textbf{G}enre-\textbf{A}ware \textbf{S}emantic \textbf{C}hange), a novel dynamic Bayesian mixture model for semantic change, where the evolution of word senses over time is based on distributional information and on additional features, specifically genre. GASC can decouple sense probabilities and genre prevalence, a critical task in the case of genre-unbalanced languages corpora, 
and can incorporate different categorical metadata, such as author, geography, or style. GASC was developed for Ancient Greek and represents the state-of-the-art in computational modelling of lexical semantic change for this language. 


On the other hand, word-embedding models have become the most common methods adopted in lexical semantic change detection \citep{kutuzov-etal-2018-diachronic} and an open question remains regarding which methods are most appropriate for ancient languages. 
In this chapter we offer the first systematic evaluation of Bayesian dynamic mixture models and word-embeddings models for semantic change in Latin and Ancient Greek.
These ancient languages provide insightful test cases of automatic lexical semantic change for several reasons. First, as in many other languages, a large number of Latin and Ancient Greek words are polysemous \citep{bakkerregister2010}, and polysemous words offers us a chance to study semantic variation, particularly across genres, and its relation to semantic change \citep{Leiwoetal2012}. 
Also, the literary traditions of these two languages have rich transcribed high-quality corpora covering a large number of literary genres. Moreover, they offer the opportunity to test the performance of different methods on use data spanning several centuries. Finally, we can rely on the scholarship of these languages to validate our computational systems. 

\begin{figure*}[ht]
\centering
\includegraphics
[width=0.75\columnwidth]{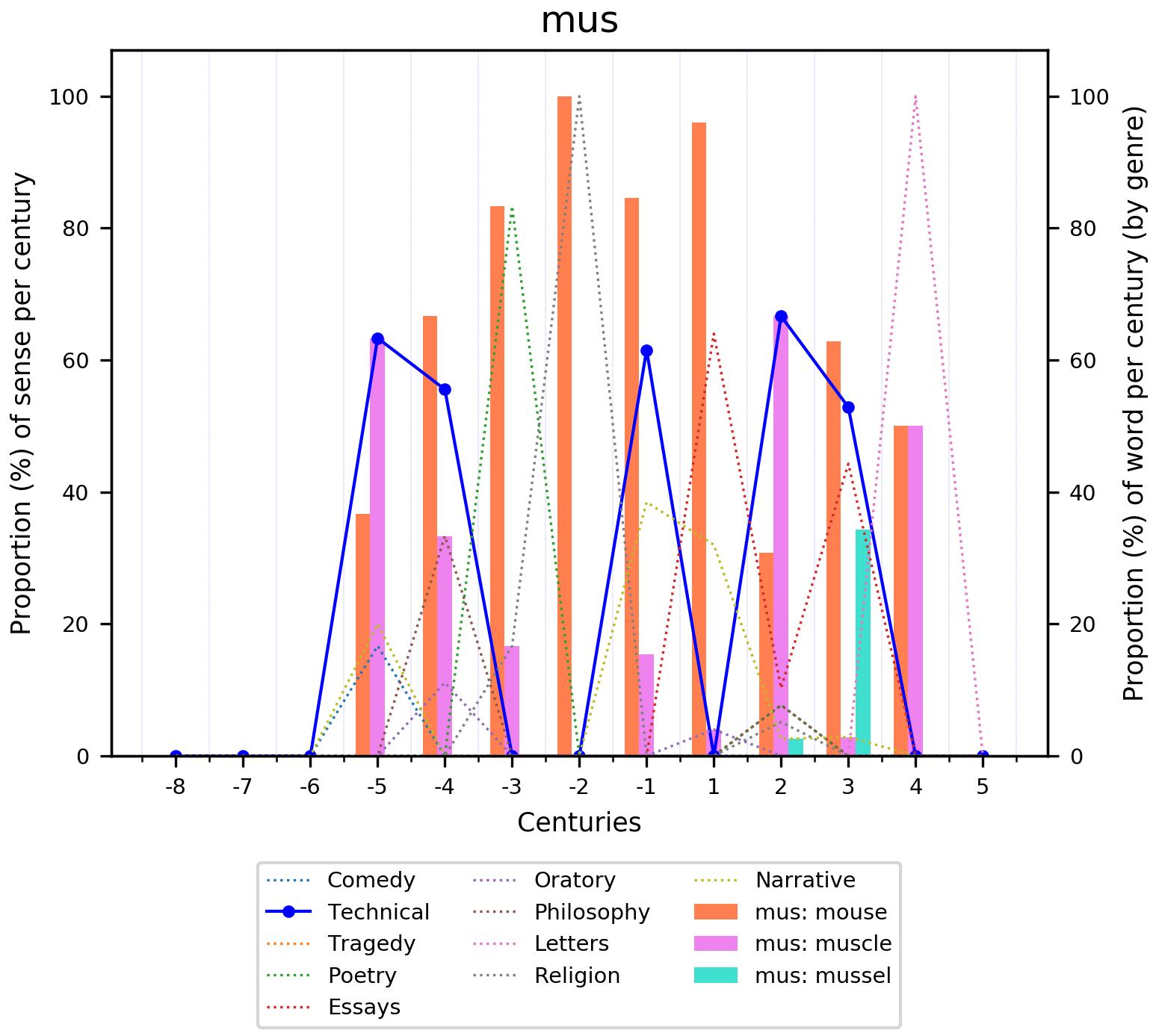}
\caption{Distribution of \emph{mus} `mouse'/`muscle'/`mussel' by genre vs its senses over time \citep{perrone-etal-2019-gasc}. Lines track \emph{mus} proportions in each genre and century, while bars show the \emph{mus} occurrence proportions with each sense and century.}
\label{musvar}
\end{figure*}

The word \emph{mus} is an example of polysemous word (it can mean `mouse', `mussel' or `muscle'). The variation in the distribution of meanings over time per genre is displayed in Figure \ref{musvar}. In this graph, lines represent the percentage of the occurrences of the target word in a literary genre across centuries, while bars represent the percentage of the occurrences of a specific sense of \emph{mus} across centuries. When the trend in any line agrees with the one for any set of bars (for instance, the distribution of `muscle' over time tracks the blue line corresponding to the the distribution of mus in technical genres), there might be evidence of genre-related changes.
In technical texts, we expect polysemous words to have a technical sense (`muscle' in the case of \emph{mus}). On the other hand, in works more closely representing general language (comedy, oratory, historiography) we expect words to appear in their more concrete and less metaphorical senses (`mouse' or `mussel' in the case of \emph{mus}), although we cannot always assume that the same distribution holds in a number of other genres, such as philosophy and tragedy.

\section{Related work}
\label{sec:related}

In recent years, NLP research has made great advances in the area of semantic change detection and modelling, with methods ranging from topic-based models \citep{boydgraber2007, cook2014novel,lau2014learning,wijaya2011understanding, frermann2016bayesian}, to graph-based models \citep{mitra2014s,mitra2015automatic,tahmasebi2017finding}, and word embeddings \citep{kim2014temporal,Basile2018,kulkarni2015statistically,hamilton2016diachronic,dubossarsky2017outta,tahmasebi2018study,rudolph2018dynamic,jatowt2018every,dubossarsky2019timeout}, to cite but a few.\footnote{For an overview of the NLP literature, we refer to
\citet{tahmasebi2018survey}, and \citet{kutuzov-etal-2018-diachronic} for a focus on neural embeddings. For an overview of the existing challenges in modelling and detecting semantic change, we refer to \citet{hengchen2021challenges}.}
However, models used in previous work are purely based on words' lexical distribution information and do not account for language variation features such as text type or genre. One reason for this is that genre-balanced corpora (such as COHA in \citealt{davies2012expanding}) or single-genre corpora (such as newspapers, or Twitter \citep[e.g.,][]{shoemark-etal-2019-room}) are typically used. 
However, the strong role played by such factors in determining the sense of a word in context has been acknowledged in NLP research at least since \citet{gale}'s idea of ``one sense per discourse'', according to which polysemous words tend to display the same sense in the same discourse. This principle has been widely adopted in word sense disambiguation research, with some more recent adaptations such as ``one sense per Wikipedia Category'' \citep{scarlini}.

Semantic change in ancient languages, especially on a large scale and over a long time period, is an under-explored research area. Previous work has mainly been  qualitative in nature, due to the complexity of the phenomenon \citep[cf. e.g.][]{Leiwoetal2012, clackson}. 
Some work has been done on training word embeddings on Ancient Greek \citep{rodda2019}) and Latin \citep{sprugnoli} corpora, but not in a diachronic perspective.
With the exception of a few works \citep{Bamman2011,eger2016linearity,Rodda2017,perrone-etal-2019-gasc,mcgillivray2019computational}, two of which this chapter is based on and completes, no previous work has focussed on ancient languages.\footnote{To this list, we add the very recent SemEval 2020 Task 1 shared task on Unsupervised Lexical Semantic Change Detection \citep{schlechtweg-etal-2020-semeval}, which had Latin as one of its four target languages.}

Recent work on languages other than English is rare but exists: \citet{falk2014quenelle} use topic models to detect changes in French and \citet{hengchen2017does} uses similar methods to tackle Dutch. 
\citet{cavallin2012automatic} and \citet{tahmasebi2018study} focus on Swedish, with the comparison of verb-object pairs and word embeddings, respectively.
\citet{zampieri2016modeling} use SVMs to assign a time period to text snippets in Portuguese, and \citet{tang2016semantic} work on Chinese newspapers using S-shaped models. 
Most work in this area focusses on simply detecting the occurrence of semantic change, while  \citet{frermann2016bayesian}'s system, SCAN, takes into account synchronic polysemy and models how the different word senses evolve across time. 
More recently French has been further tackled by \citet{jawahar-seddah-2019-contextualized}, \citet{frossard-etal-2020-dataset} and \citet{montariol-allauzen-2020-etude}, and German has been the focus of extensive work \citep{schlechtweg-etal-2017-german,schlechtweg2018durel,schlechtweg-etal-2019-wind,schlechtweg-etal-2020-semeval}.

Our work bears important connections with the topic model literature. The idea of enriching topic models with document-specific author meta-data was explored in \citet{Rosen-Zvi2004} for the static case. Several time-dependent extensions of Bayesian topic models have been developed, with a number of parametric and nonparametric approaches \citep{bleilafferti, gamma, xing, topictime, perrone2017}. In this chapter, we transfer such ideas to semantic change, where each datapoint is a bag of words associated to a single sense (rather than a mixture of topics). Excluding cases of intentional ambiguity, which we expect to be rare, we assume that there are generally no ambiguities in a context, and each word instance maps to a single sense. We acknowledge that this assumption can be seen as going against historical semantics literature (e.g. \citealt{traugott2001regularity}) which states that variation in context is the seed of semantic change.


\section{The corpora}\label{sec:data}

In order to conduct our experiments, we made use of two large diachronic corpora of Latin and Ancient Greek: LatinISE \citep{mcgillivray-kilgarriff} for Latin and the Diorisis Annotated Ancient Greek Corpus \citep{diorisis2018} for Ancient Greek. Our models require genre information.  
Genre-annotated corpora are not particularly common in NLP, where most tasks rely on specific genres (e.g. Twitter) or on genre-balanced corpora such as COHA \citep{davies2002corpus}, but they are more prevailing within the humanities, and especially classics. 
Additionally, research on automated genre identification has been flourishing for decades (e.g. \citealt{kessler1997automatic}), making the need for genre information in a potential corpus not as much of a hindrance as it can be thought.\footnote{While the influence of genre has been extensively studied in historical linguistics (see, for example, the extensive work by \citealt{biber1989drift}), we use in this chapter a slightly different notion of `genre': literary genre, as defined by classicists.} 

The Diorisis Annotated Ancient Greek Corpus 
contains 820 texts spanning between the beginnings of the Ancient Greek literary tradition (8\textsuperscript{th} century BCE) and the 5\textsuperscript{th} century CE. It is  lemmatized and part-of-speech-tagged and contains 10,206,421 word tokens. Diorisis is the largest openly available annotated corpus of Ancient Greek. The corpus covers a number of Ancient Greek literary and technical genres: poetry (narrative, choral, epigrams, didactic), drama (tragedy, comedy), oratory, philosophy, essays, narrative (historiography, biography, mythography, novels), geography, religious texts (hymns, Jewish and Christian Scriptures, theology, homilies), technical literature (medicine, mathematics, natural science, tactics, astronomy, horsemanship, hunting, politics, art history, rhetoric, literary criticism, grammar), and letters. 

The LatinISE corpus \citep{mcgillivray-kilgarriff} covers 1,274 texts from between the beginnings of the Latin literary tradition (2\textsuperscript{nd} century BCE) and the contemporary era (21\textsuperscript{st} century CE). It has been automatically lemmatized and part-of-speech tagged. 
A domain expert manually added genre information for the following genres: comedy, essays, law, letters, narrative, oratory, philosophy, poetry, christian, technical, tragedy. All Christian writings (including letters and poems) were assigned the genre Christian, this excludes philosophical but not theological or ecclesiological treatises composed by Christian writers. 

\section{Bayesian semantic change models}

\subsection{Domain knowledge elicitation}\label{sec:elicitation}
While NLP provides powerful tools to analyse texts, a central challenge is to ensure that outputs are explainable and that new discoveries can be placed within the context of current art in specific disciplines where NLP methods are applied. Bayesian methods have proved very useful within scientific modelling to incorporate domain explanations. In the Bayesian setting, expert judgements can be embedded directly into a probabilistic framework in the form of a prior. For instance, if historians know that a certain sense was popular in a given century, this information can be directly encoded into the model by changing the prior probability distribution for that sense. Data can then be analysed from these belief statements and a prior to posterior analysis performed, which helps domain experts adjust their beliefs in the light of the new available information (see for
example \citealt{Smith2010, Hagan2014}). These new outputs will be consistent with the explanations embedded within the probabilistic model, making results interpretable.

The challenge of applying Bayesian reasoning within the humanities is that typically domain experts have not been trained to reason probabilistically. Therefore, it is not possible to ask domain experts to provide direct probabilistic inputs to the Bayesian model. What it is possible instead is to elicit structural information, which can take a wide range of forms depending on the domain \citep{Wilk2019}. These structural models can usually be represented by a graph (i.e., as a set of nodes and connecting arcs) which capture the fundamental entities and their relationships. For example, an expert may know that a certain author predominantly uses \emph{mus} to mean 'mouse'. The Bayesian modeller can then simply introduce a new node representing the author and condition the probability of using senses to the author variable. Once the graphical model is in place, we let the data quantify a joint probability model. 

This work leverages the Bayesian network, one of the most developed structural models of this type. This structure embeds simple assertions about what measurements might be informative, in a way described in \citet{Korb2009}, \citet{Smith2010}, and \citet{Pearl2014}. Working backwards from the properties of the object of interest, we produce sequentially a collection of direct and indirect influences across the whole domain. The composite of the relationships can then be expressed by a single graph, called a plate diagram (see Figure~\ref{diagram} for such a plate diagram of our model). This plate diagram determines the factorisation of the corresponding probability density over these measurements.

We aimed to apply these structural elicitation techniques to study semantic variation and change in Latin and Ancient Greek. From discussions with Ancient Greek and Latin experts who have extensive experience with the corpora at hand, it emerged that one of the main drivers of this variation was the particular genre of the text. For instance, in works more closely related to general language (i.e. non-specialised, or non purely poetic language), such as comedy or historiography, we expect words to appear in their concrete and less metaphorical senses. The Ancient Greek word \emph{mus} within a technical text would more likely mean `muscle', while in narrative texts the meaning would more likely be `mouse'. Such variations were believed to abound within the studied corpus. Since both the genre of texts was known to vary over time and text preservation to the current date depended on genre, any analysis which ignored genre might deduce a spurious change in overall meaning simply explainable from drifts in genre and selection effects influencing preservation. Having elicited this domain judgement, it was clear how to proceed. We simply modified the structure of our Bayesian model by adding genre as an additional observable variable (or node in
the plate diagram). Conditioning on the observed genre, we could then have a specific distribution over senses accounting for genre-specific word usage patterns. Details of the model are given in the next section. 

\subsection{Genre-aware semantic change}\label{sec:model}

\begin{figure*}[ht]
\centering
\includegraphics
[width=0.9\columnwidth]{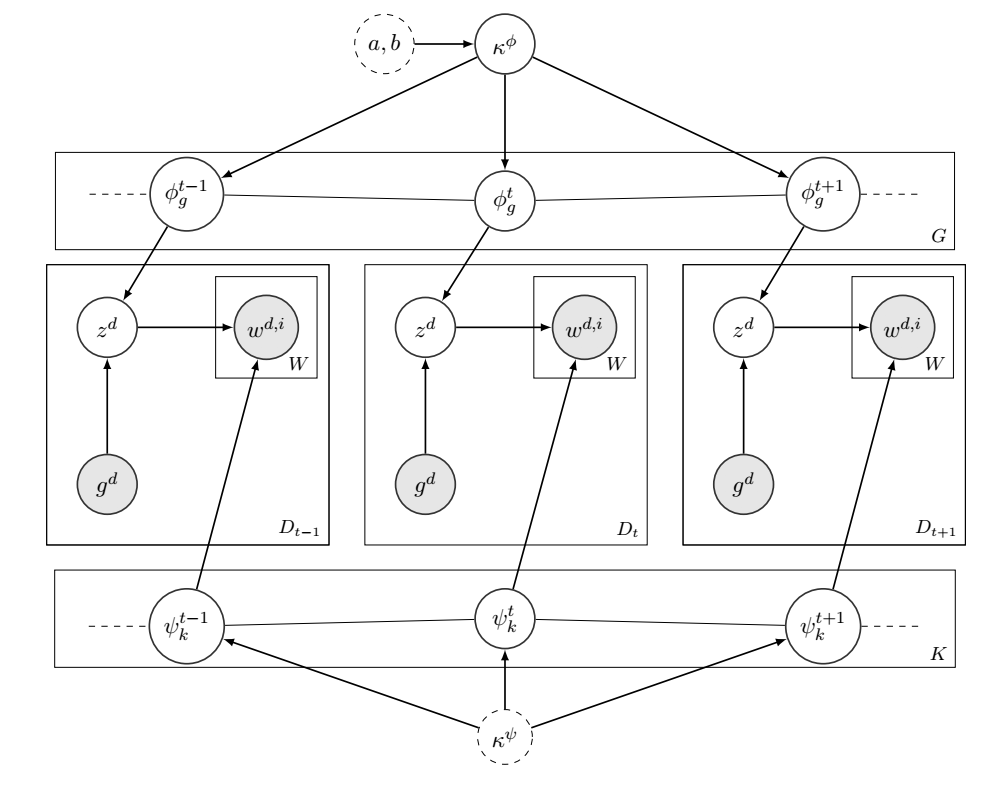}
\caption{GASC plate diagram with three time periods \citep{perrone-etal-2019-gasc}.}
\label{diagram}
\end{figure*}

A successful approach to model semantic change in Ancient Greek is GASC \citep{perrone-etal-2019-gasc}. The starting point is a lemmatised corpus pre-processed into a set of text snippets of size $W$, each containing an instance of the word under study (referred to as ``target word'' in the remainder). 
The inferential task is to detect the sense associated to the target word in the given context, and describe the evolution of sense proportions over time.


We briefly summarise the generative model for GASC (illustrated by the plate diagram in Figure~\ref{diagram}), which extends SCAN \citep{frermann2016bayesian} to be genre-aware and is described in detail in \citet{perrone-etal-2019-gasc}. First, suppose that throughout the corpus the target word is used with $K$ different senses, where we define a sense at time $t$ as a distribution $\psi^t_k$ over words from the dictionary. This statistical definition of sense is necessary to formalize the generative models presented in this work, and will be used throughout the rest of the paper.\footnote{We follow the terminology adopted by \citet{frermann} and represent the meaning of a word as a set of senses, each of which captures ``an internally coherent aspect of its meaning, and is characterized through a set of words that are associated with that sense'' \cite[173]{frermann}. We also assume that each instance of a target word in the corpus refers to one and only one sense.}
These distributions are used to generate text snippets by drawing each of their words from the dictionary based on a multinomial distribution. Based on the intuition that each genre is more or less likely to feature a given sense, we assume that each of $G$ possible text genres determines a different distribution over senses. Each observed document snippet is then associated with a genre-specific distribution over senses $\phi_{g{^d}}^t$ at time $t$, where $g{^d}$ is the observed genre for document $d$. Conditioning on the observed genre yields a specific distribution over senses accounting for genre-specific word usage patterns. 
Word and sense distributions evolve over time with Gaussian changes, allowing for smooth transitions. 


The model can be applied to different inferential goals: we can focus on the evolution of sense probabilities or on the changes within each sense. As we define a sense at time $t$ as a probability distribution over words from the dictionary, this means that we can either choose to focus on the change of the sense probability over time or on the change in probability of the words characterising that sense. For each of these aims, we can use several hyperparameter combinations for $K^\phi$, which is drawn from the prior distribution as determined by $a$ and $b$, and $K^\psi$.
To effectively detect semantic change points, the sense probabilities should not vary too smoothly over time and the bag of words should remain stable throughout the time periods.\footnote{We acknowledge that the task of detecting change points is a drastically reduced view of semantic change. Nonetheless, as further explained in Section~\ref{sec:evaluation}, this is required for ground truth evaluation.} For these reasons, we set the hyperparameters $a = b = 1$, $K^\psi = 100$ (equivalent to Setting 3 in \citealt{perrone-etal-2019-gasc}). In particular, the hyperparameter $K^\psi$ controls the homogeneity of the bag of words within the same sense and allows the emergence of new senses. This hyperparameter setting is used for SCAN and GASC on Latin, as well as for SCAN on Ancient Greek. For GASC on Ancient Greek, where the corpus size and the number of occurrences of target words is split between genres, the set of hyperparameters used is $a = 7, b = 3$, $K^\psi = 10$, as in \citet{frermann2016bayesian}.

Further quantities to be set before running the Bayesian models are the number of iterations and the window size parameter $W$. The first runs of the Bayesian models usually show high variability in the results before convergence occurs; therefore, it is necessary to use a large number of iterations, especially for small sample sizes. For posterior inference we discard the first 100 iterations (burn-in period) and we run 2,500 iterations for models on Latin and 10,000 for models on Ancient Greek. The window size parameter $W$, namely the number of words to the left and right of an instance of the target, must also be carefully chosen not to introduce noisy irrelevant contextual words. Following \citet{frermann2016bayesian} and \citet{perrone-etal-2019-gasc}, we fix the window size $W$ to 5 for all methods and languages.

\subsubsection{Posterior inference}
For posterior inference, we extend the blocked Gibbs sampler proposed in \citet{frermann2016bayesian}. The full conditional is available for the snippet-sense assignment, while  to  sample  the  sense and word distributions we adopt the auxiliary variable approach from \citet{Mimno2008}. The sense precision parameters are drawn from their conjugate Gamma priors. For the distribution over genres we proceed as follows. First, sample the distribution over senses $\phi^t_g$ for each genre $g=1,\dots,G$ following \citet{Mimno2008}. Then, sample the sense assignment conditioned on the observed genre from its full conditional:
$p(z^d \mid g^d, \textbf{w}, t, \phi, \psi) \propto p(z^d \mid g^d, t) p( \textbf{w} \mid t, z^d) = \phi^t_{g} \prod_{w \in \textbf{w}} \psi^{t,z^d}_w.$
This  setting easily extends to sample genre assignments for tasks where, for example, some genre metadata are missing.

\section{Embedding-based models}
\label{sec:embedding}

Neural-based word vectors are currently the most used representations in LSC.
While skip-gram with negative sampling \citep[SGNS,][]{mikolov2013efficient} type embeddings have the limitation that they conflate senses of a word to a single vector representation, they currently perform better than other approaches, including contextual models such as ELMO \citep{peters-etal-2018-deep} and BERT \citep{devlin2018bert}, as reported by \citet{schlechtweg-etal-2020-semeval}.

In this chapter, we compare GASC and SCAN to the current state of the art in LSC \citep[Temporal Referencing (TR),][]{dubossarsky2019timeout}, as well as with the oft-used combination of independently-trained SGNS models that are subsequently aligned using Orthogonal Procrustes (OP) proposed by \citet{hamilton2016diachronic}.
Both models are very similar and rely on the same algorithm with the difference that TR, in which target words have different representations for every time bin but context words do not, has repeatedly been shown to produce much less noisy models \citep[e.g. in][]{cassotti-etal-2020-gmctsc,zamora-reina-etal-2020-dcc-uchile}.
In order to compare their performance with GASC, we train models on the whole corpus (``NAIVE''), as well as on genre subcorpora. For Ancient Greek we train models on Technical, NOT-technical, Narrative, NOT-narrative subcorpora, while Latin is divided between Christian and NOT-Christian.

\section{Evaluation} 
\label{sec:evaluation}
Evaluating models tackling lexical semantic change is notoriously challenging. 
\cite{schlechtweg-etal-2020-semeval} present the first shared task on unsupervised lexical semantic change detection, organized as part of the SemEval 2020 workshop. The task focusses on two subtasks:  a binary classification task (for a set of target words, decide which words lost or gained senses
between a time period $t_{1}$ and a time period $t_{2}$) and a ranking subtask (rank the same set of target words according to their degree of lexical semantic change between $t_{1}$ and $t_{2}$).
The task provides gold standard sets for three extant languages (English, German, and Swedish) and one extinct language (Latin).

The Latin gold standard reflects the lexical semantic change in a portion of the Latin lexicon from the pre-Christian (BCE) era and the Christian (CE) era. For each of 40 lemmas selected from the corpus, expert annotators annotated 30 sentences extracted from a subcorpus of LatinISE consisting of texts from the BCE era, and 30 sentences from the CE era. For each sentence, the annotators selected one of four values (4 -- Identical, 3 -- Closely Related, 2 -- Distantly Related, 1 -- Unrelated) for each dictionary sense of the lemma, indicating the degree of similarity between the usage of the lemma reflected in the sentence and the dictionary sense. This choice of design implying that every target word has a closed set of possible senses corresponding to those listed in their respective dictionary entries is justified in the original paper.

The annotated data was analysed with a clustering technique that identified 26 lemmas as ``changed'' lemmas (meaning that they underwent lexical semantic change between the BCE and the CE era) and 14 lemmas as ``unchanged'' (meaning that they did not undergo lexical semantic change). For details on the clustering and the annotation, see \cite{schlechtweg-etal-2020-semeval}. The SemEval task competition and the subsequent article describing a subset of the systems that took part in it offers the first systematic evaluation of state-of-the-art systems for automatic lexical semantic change detection.

\subsection{Ground truth evaluation}

Word embedding models build vector representations of a word for every time slice at hand. For two time intervals $t_1$ and $t_2$, we then use a similarity measure (usually, cosine similarity) as a proxy to determine the semantic change between the vectors $w_{t_1}$ and $w_{t_2}$ for a specific word between these time slices:

$$cosine\_similarity(w_{t_1}, w_{t_2}) =
\frac{w_{t_1} \cdot w_{t_2}}{\|  w_{t_1} \| \| w_{t_2} \|},$$ 
where $\|  \cdot \|$ denotes the Euclidian norm. A high cosine similarity (e.g., close to $1$) means no difference for word $w$ between time slices $t_1$ and $t_2$, and a low cosine similarity indicates a high difference.\\  
As our ground truth consists of a binary classification (no-change / change, cf. Section~\ref{sec:results}), we must transform the cosine similarity value, bounded between -1 and 1, into a decision. 
While manual thresholding on the cosine is usually applied, recent work \citep{zhou-etal-2020-temporalteller} shows that determining the threshold in a data-driven way is beneficial. We thus follow prior work on Latin and fit a Gamma distribution of the cosine similarities for all target words between $t_0$ and $t_1$, and consider every cosine similarity below the 75-quantile value as the threshold for a change decision.\footnote{We thank Jinan Zhou and Jiaxin Li for providing us with their implementation.}

On the other hand, dynamic Bayesian mixture models, such as SCAN and GASC, are designed to infer the smooth evolution of sense probabilities over time. We adapt these methods to detect sense change points as follows. First, we compute the mean and standard deviations of the posterior sense probabilities over time based on the Gibbs samples obtained during inference. Then, we infer that there has been a significant drop or rise of a sense if its posterior mean probability changes by at least two standard deviations over time. In case of a significant drop we infer that a sense disappeared, and in case of a significant increase we infer that a new sense appeared in the data. If sense probabilities do not change significantly over time, we conclude that no meaning change occurred. Note that, unlike SCAN, GASC outputs a sense probability over time for each genre, and we thus check across all genres whether a significant change of sense probability occurred over time. While we adopt this approach for simplicity, change point analysis has been studied extensively in the context of Gaussian dynamic state space models. We refer to \citet{west1997} and \citet{FRUE2006} for more sophisticated approaches to detect change points, which also allow for returning a probability distribution over change points.

\section{Experiments}
\label{sec:results}

\subsection{Tasks and baselines}
We compared SCAN and GASC to a wide range of baselines on the task of detecting binary change in both Latin and Ancient Greek. \cite{perrone-etal-2019-gasc} and \cite{annodata} present a gold standard set created for the purpose of evaluating GASC on Ancient Greek. This set consists in the sense annotation of corpus sentences for three words (\emph{mus} `mouse'/ `muscle'/ `mussel', \emph{harmonia} `fastening'/ `agreement'/ `musical scale, melody', \emph{kosmos} `order'/ `world'/ `decoration'). These lemmas display a high degree of clear-cut polysemy,\footnote{By clear-cut polysemy, we mean that the different senses of a word are not strongly related.} especially across genres \citep{Liddell:1996, pollittancient1974}, and were chosen as ``non-changed'' words. We considered two additional lemmas, which display a degree of lexical semantic change in the time period under study, \emph{parabole} `comparison'/`parable' and \emph{paradeisos} `garden'/`paradise' \citep{mcgillivray2019computational}. \emph{Paradeisos} is an Avestan loan word that first appeared in Greek in the fifth century BCE to indicate a `royal park' and probably became common after the Macedonian conquest of the Persian empire. This word was chosen by the Greek translators of the Pentateuch to refer to the garden of Eden around the third century BCE \citep{christidis2007}. The meaning of  \emph{parabole}, in turn, specialized from that of `comparison' to that of `short moral narrative' with the New Testament (first century CE). For Latin, we made use of the SemEval task's gold standard, consisting of 26 ``changed'' lemmas and 14 ``non-changed'' lemmas between the BCE era and the CE era. We start by visualizing the smooth semantic change inferred by GASC, and then compare the ability of dynamic Bayesian mixture models to detect binary semantic change with the state of the art, both on Latin and Ancient Greek.

\subsection{Smooth semantic change}
\begin{figure*}[ht]
\centering
\includegraphics[width = \textwidth]{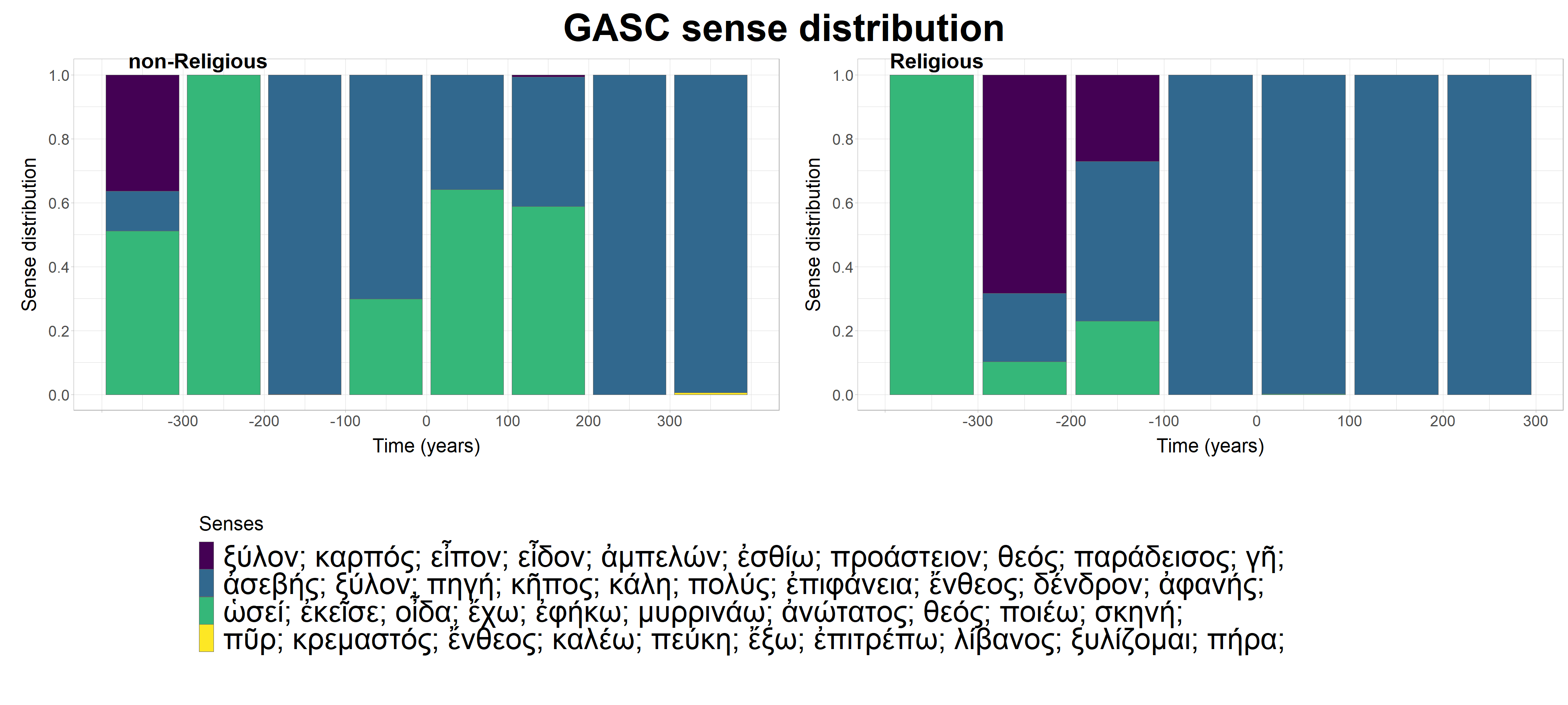}.
\caption{Semantic change in Ancient Greek. Visualization of the probability distributions produced by GASC on the Religious genre for the word \emph{paradeisos} `garden' / `paradise'. Negative numbers refer to years BCE.}
\label{paradeisos}
\end{figure*}

Dynamic Bayesian mixture models are able to infer the full evolution of sense probabilities over time. In particular, GASC is able to do so for each genre provided as input. Figure \ref{paradeisos} shows the time distribution of the senses of \emph{paradeisos} outputted by GASC run on Religious vs. non-Religious genre. 
The four senses identified by GASC may be interpreted as identifying the meaning `garden' (senses 3-green and 4-yellow),  and `garden of Eden/(Biblical) paradise' (senses 1-purple and 2-blue). The two senses are not easily distinguishable (since the Biblical paradise is described as a physical garden) and all senses share a number of words, including, notably, \emph{theos} `God' or the derived adjective \emph{entheos} `inspired by God'. However, the first two senses contain a number of words that are easily identifiable as connected to the Biblical narration of the fall of man (e.g. \emph{karpos}, (the forbidden) `fruit' and \emph{esthio}, `eat') while the remaining senses suggest references to other proverbial gardens (e.g. \emph{kremastos} `hanging' garden of Babylon). The diachronic evolution of sense distributions in the plots shows that the Biblical meaning comes to rise around the third century BCE in religious texts, which corresponds precisely to the beginning of the translation of the Bible in Greek, and will prevail throughout the Christian era. The graph displaying the computed distribution of senses in non-religious genres captures well the fact that between the first century BCE and the second century CE \emph{paradeisos} is attested a number of times in the works of historians and geographers represented in the corpus. After the third century, this word is very rarely attested in the works included in the Diorisis corpus and almost half of its occurrences in non-religious texts refer to the Biblical garden of Eden.

\subsection{Binary semantic change}
\label{binary-eval}
Next, we evaluated the ability to recover ground truth about binary semantic change on both Latin and Ancient Greek. For Latin, we recall that ground truth consists of 40 target lemmas, 26 of which underwent semantic change. We ran the genre-aware baselines by specifying whether a text belongs to the Christian genre or not. Results in Table~\ref{table:binary_change} show that Bayesian models are highly competitive with the best baseline obtained in the SemEval task, with SCAN achieving the highest F1 score. This is striking as dynamic Bayesian mixture models are designed for capturing smooth semantic change over time, rather than binary semantic change across a pair of time points. In addition, only focusing on non-Christian genres decreases the recall of SGNS and TR. This is expected as the 26 lemmas that underwent semantic change did so due to the rise of a new Christian meaning. 

We then evaluated each method on Ancient Greek, further adapting SGNS and TR to use genre information and focus on technical and narrative texts. To evaluate GASC, we use Religious as the genre for \emph{parabole} and \emph{paradeisos}, while Technical and Narrative for \emph{mus}, \emph{harmonia} and \emph{kosmos}, with results being averaged across the five words. Results are shown in Table~\ref{table:binary_changeAG}. While the small number of target words makes these results mainly illustrative, dynamic Bayesian mixture models emerge as competitive approaches. Consistently with Latin, GASC and SCAN outperform most baselines.
To better understand how differently SCAN and TR (the two best-perfoming systems) behave, we refer to the confusion matrix in Table~\ref{tab:confusion-matrix}.

\begin{table*}[h!]
\centering
\footnotesize

\begin{tabular}{lccc}
\hline
\multicolumn{1}{l}{Latin (BCE/CE)} & \multicolumn{1}{c}{Precision} & \multicolumn{1}{c}{Recall} & \multicolumn{1}{c}{F1 score} \\ \hline
\emph{SCAN}                  & \multicolumn{1}{c}{0.684}  & \multicolumn{1}{c}{1.000}  & \multicolumn{1}{c}{\textbf{0.813}}   \\
\emph{GASC}               & \multicolumn{1}{c}{0.650}  & \multicolumn{1}{c}{0.920}  & \multicolumn{1}{c}{0.762}  \\
\emph{SGNS\_NOT-christian}& \multicolumn{1}{c}{1.000}  & \multicolumn{1}{c}{0.308}  & \multicolumn{1}{c}{0.471}  \\
\emph{SGNS\_NAIVE}& \multicolumn{1}{c}{0.900}  & \multicolumn{1}{c}{0.347}  & \multicolumn{1}{c}{0.500}  \\
\emph{TR\_NOT-christian}& \multicolumn{1}{c}{0.667}  & \multicolumn{1}{c}{0.231}  & \multicolumn{1}{c}{0.343}  \\
\emph{TR\_NAIVE}& \multicolumn{1}{c}{0.769}  & \multicolumn{1}{c}{0.769}  & \multicolumn{1}{c}{0.769}  \\
\emph{Best Baseline}                 & \multicolumn{1}{c}{0.650}  & \multicolumn{1}{c}{1.000}   & \multicolumn{1}{c}{0.788}  \\ \hline
\end{tabular}
\caption{Semantic change in Latin. Comparison of SCAN and GASC with SGNS, TR and the best baseline from the SemEval task. Results in terms of precision, recall, and F1-score (``F1'') averaged across all 40 available words. Results for TR\_NAIVE are by \citet{zhou-etal-2020-temporalteller}.
}
\label{table:binary_change}
\end{table*}

\begin{table}[]
\centering
\footnotesize
\begin{tabular}{lcccc}
\hline
System     & TP & TN & FP & FN \\ \hline
\emph{SCAN} & 26 & 2  & 12 & 0  \\ 
\emph{TR\_NAIVE}   & 20 & 8  & 6  & 6  \\ \hline
\end{tabular}
\caption{\small Confusion matrix for binary change in Latin for SCAN and Temporal Referencing. TP = true positive, TN = true negative, FP = false positive, FN = false negative.}
\label{tab:confusion-matrix}
\end{table}

\begin{table*}[h!]
\centering
\footnotesize

\begin{tabular}{lccc}
\hline
\multicolumn{1}{l}{Ancient Greek } & \multicolumn{1}{c}{Precision} & \multicolumn{1}{c}{Recall} & \multicolumn{1}{c}{F1 score} \\ \hline
\emph{GASC}& \multicolumn{1}{c}{0.600}  &
\multicolumn{1}{c}{1.0}  & \multicolumn{1}{c}{\textbf{0.750}}  \\
\emph{SCAN}& \multicolumn{1}{c}{0.500}  &
\multicolumn{1}{c}{0.667}  & \multicolumn{1}{c}{0.571}  \\
\emph{SGNS\_NOT-technical}& \multicolumn{1}{c}{0.333}  & \multicolumn{1}{c}{0.500}  & \multicolumn{1}{c}{0.400}  \\  
\emph{SGNS\_NOT-narrative}& \multicolumn{1}{c}{0.333}  & \multicolumn{1}{c}{0.500}  & \multicolumn{1}{c}{0.400} \\ 
\emph{SGNS\_technical}& \multicolumn{1}{c}{0.000}  & \multicolumn{1}{c}{0.000}  & \multicolumn{1}{c}{0.000}  \\ 
\emph{SGNS\_narrative}& \multicolumn{1}{c}{0.000}  & \multicolumn{1}{c}{0.000}  & \multicolumn{1}{c}{0.000} \\ 
\emph{SGNS\_NAIVE}& \multicolumn{1}{c}{0.333}  & \multicolumn{1}{c}{0.500}  & \multicolumn{1}{c}{0.400} \\ 
\emph{TR\_NOT-technical}& \multicolumn{1}{c}{0.400}  & \multicolumn{1}{c}{1.0}  & \multicolumn{1}{c}{0.571} \\ 
\emph{TR\_NOT-narrative}& \multicolumn{1}{c}{0.333}  & \multicolumn{1}{c}{0.500}  & \multicolumn{1}{c}{0.400} \\ 
\emph{TR\_technical}& \multicolumn{1}{c}{0.000}  & \multicolumn{1}{c}{0.000}  & \multicolumn{1}{c}{0.000}  \\ 
\emph{TR\_narrative}& \multicolumn{1}{c}{0.500}  & \multicolumn{1}{c}{1.0}  & \multicolumn{1}{c}{0.667}  \\ 
\emph{TR\_NAIVE}& \multicolumn{1}{c}{0.333}  & \multicolumn{1}{c}{0.500}  & \multicolumn{1}{c}{0.400} \\ \hline 
\end{tabular}
\caption{Semantic change in Ancient Greek. Comparison of SGNS, TR, GASC and SCAN on the task of detecting binary semantic change. Results in terms of precision, recall, and F1-score (``F1'') are averaged across the 5 available words.
}
\label{table:binary_changeAG}
\end{table*}

\section{Discussion and conclusion}\label{sec:discussion}
This work investigates semantic change in Latin and Ancient Greek through several state-of-the-art models. We adapted, discussed and applied a number of algorithms to the case of ancient languages. 
The adoption of quantitative corpus-based approaches in historical linguistics is growing \citep{jenset}. However, computational approaches to lexical semantic change detection have not yet been widely used in historical linguistics research  \citep{mcgillivray2020}, although a few steps in this direction have been taken (see e.g. \citealt{keersmaekers,rodda2019,mcgillivray2019computational}).
In spite of their limited use in lexical semantic change detection,  dynamic Bayesian mixture models allow practitioners to embed domain expert knowledge and provide interpretable outputs. 

We provided a systematic comparison of SCAN and GASC, two recent models from this family, with state-of-the-art embedding-based models, such as SGNS and Temporal Referencing. In addition, we transformed embedding models to account for genre information and provided a new evaluation framework to detect binary semantic change based on expert-annotated data. 

Our experiments show that Bayesian models are highly competitive at detecting binary change, beating all baselines on Ancient Greek and Latin. These results, together with the ability to provide full representations of the evolution of word senses, indicate Bayesian dynamic mixture models as successful approaches to study semantic change in ancient languages.


This work can also be seen as a step towards the development of richer evaluation schemes and models that can embed expert judgement. We have shown how including genre can improve the understanding of the historical development of words in a corpus. 
We argue that the next process to be captured from semantic change models is the archiving of historical texts. The entirety of the relevant documents extant at any time in history is an obvious reference population against which we perform inference. While any analysis based on a currently extant corpus could be biased, Bayesian models embedding historical domain knowledge enable us to de-bias the study (e.g., by accounting for missing texts when inferring the popularity of a sense).
There are essentially three different necessary conditions for a text to be extant at any given time. The first is the decision of a librarian to add a particular document to a library, the second is whether or not that text is preserved or destroyed during the passage of time, and the third is the inability of researchers to access documents extant at the current time. A Bayesian analysis enables us to embed a probabilistic description of such a development. For example, many texts within a given corpus will have their own associated provenance, which can be used to help inform the nature of the likely extant corpus. This allows historical insights and extra data to be drawn into the analysis and better inform historical conjectures.  The explicit development of such models is ongoing, and we will report our findings in future work.

\setcounter{secnumdepth}{0}

\section{Acknowledgements}

This work was supported by The Alan Turing Institute under the EPSRC grant EP/N510129/1 and the seed funding grant SF042. 
SH's work is funded by the project \textit{Towards Computational Lexical Semantic Change Detection} supported  by the Swedish Research Council (2019--2022; dnr 2018-01184).




\bibliography{main.bib}
\bibliographystyle{acl_natbib}

\end{document}